\documentclass[10pt,twocolumn,twoside]{IEEEtran}
\usepackage{cite}

\usepackage{geometry}
 \usepackage[pdftex]{graphicx}
 \graphicspath{{../pdf/}{../jpeg/}}
\usepackage{amsmath}
\usepackage{cases}

\usepackage{algorithm}
\usepackage{algpseudocode}
\usepackage[font=small, skip = 2pt]{caption}
\usepackage{array}
\usepackage{diagbox}
\usepackage{multirow}
\newcommand{\subparagraph}{}
\usepackage{titlesec}

\usepackage[caption=false,font=footnotesize]{subfig}
\usepackage{layouts}

\usepackage{float}
\usepackage{url}

\titlespacing{\section}{0pt}{1.3ex plus 0.2ex minus .2ex}{0.6ex plus .0ex}
\titlespacing{\subsection}{0pt}{0.7ex plus 0.2ex minus .2ex}{0.5ex plus .0ex}
\DeclareCaptionOption{tight}[]{%
\captionsetup{farskip=2pt,topadjust=0pt,captionskip=2pt,nearskip=2pt,margin=0pt,font=footnotesize}
}
\begin{document}

% paper title.
\title{Power Optimizations in MTJ-based Neural Networks through Stochastic Computing}
\author{\vspace{-2mm}\IEEEauthorblockN{Ankit Mondal and Ankur Srivastava}\\
\IEEEauthorblockA{Department of Electrical and Computer Engineering, 
University of Maryland, College Park, MD 20783, USA}\\
Email: $\{amondal2, ankurs\}@umd.edu$\vspace{-6mm}}
%copyright
%\IEEEpubid{978-1-5090-6023-8/17/\$31.00 \copyright 2017 IEEE}
% make the title area
\maketitle
\begin{abstract}
Artificial Neural Networks (ANNs) have found widespread applications in tasks such as pattern recognition and image classification. However, hardware implementations of ANNs using conventional binary arithmetic units are computationally expensive, energy-intensive and have large area overheads. Stochastic Computing (SC) is an emerging paradigm which replaces these conventional units with simple logic circuits and is particularly suitable for fault-tolerant applications. Spintronic devices, such as Magnetic Tunnel Junctions (MTJs), are capable of replacing CMOS in memory and logic circuits.   In this work, we propose an energy-efficient use of MTJs, which exhibit probabilistic switching behavior, as Stochastic Number Generators (SNGs), which forms the basis of our NN implementation in the SC domain. Further, error resilient target applications of NNs allow us to introduce Approximate Computing, a framework wherein accuracy of computations is traded-off for substantial reductions in power consumption. We propose approximating the synaptic weights in our MTJ-based NN implementation, in ways brought about by properties of our MTJ-SNG, to achieve energy-efficiency. We design an algorithm that can perform such approximations within a given error tolerance in a single-layer NN in an optimal way owing to the convexity of the problem formulation. We then use this algorithm and develop a heuristic approach for approximating multi-layer NNs. To give a perspective of the effectiveness of our approach, a 43\% reduction in power consumption was obtained with less than 1\% accuracy loss on a standard classification problem, with 26\% being brought about by the proposed algorithm.
\end{abstract}
\vspace{-4mm}
\section{Introduction}
The capability of the human brain to learn and solve complex problems have inspired advancements in areas of neuroscience, 
artificial intelligence and machine learning. Decades of research in Artificial Neural Networks (ANNs), despite our limited understanding of 
biological Neural Networks (NNs), have shown promising results in applications such as pattern recognition and image classification \cite{samarasinghe2016neural}. 
However, a typical ANN can have thousands of neurons and synapses, making their hardware implementation both computation and 
memory-intensive. This has prompted the development of optimization techniques at different levels of these complex networks to 
achieve energy efficiency \cite{venkataramani2014axnn} \cite{mrazek2016design}.

Approximate Computing is an emerging concept which involves the computation of imprecise results in order to achieve significant 
reductions in power consumption \cite{han2013approximate}. The inherent error-resilience of Recognition, Mining and Synthesis applications make them a 
perfect candidate for such trade-off between the quality of results and the energy requirements. A similar paradigm is Stochastic Computing 
(SC) which concerns the use of low-cost logic gates, instead of binary arithmetic units, for computations \cite{alaghi2013survey}.
%It exploits the probabilistic nature of computations and uncertain circuit behavior, such as those arising from process variations in modern technology . 
%It aims to exploit the probabilistic nature of certain computations arising due to manufacturing process variations or soft errors in modern circuits. 
In SC, data, which are interpreted as probabilities and called Stochastic Numbers (SNs), are represented in the form of bit streams of 0s and 1s and generated by circuits called Stochastic Number Generators (SNGs). SC has been shown to be significantly energy-efficient when compared to conventional methods.

Magnetic Tunnel Junction (MTJ) is one of several emerging spintronic devices.  Apart from non-volatility, its high integration density, 
scalability and CMOS compatibility make it a suitable candidate for replacing CMOS in future memory devices \cite{kim2015technology}. The Spin-Transfer Torque RAM, which is based on 
MTJs, has been explored as a memory device. While a lot of research has 
focused on reducing its critical switching current density to lower the write energy, attempts have been made to exploit the probabilistic switching characterics of MTJs to use them as SNGs, such as in \cite{de2015stochastic}, that could produce bit streams representing any fraction between 0 and 1.

This paper integrates SC based on MTJs into ANNs and explores the different ways of achieving energy efficiency at both the device level and the network level, in the latter through approximations. Our contributions are summarized as follows:
\begin{itemize}
\item We outline the characteristics of an MTJ with regard to switching time and energy, develop a low-power MTJ-SNG by exploiting the properties of SC, and compare it with the baseline.
\item We propose the use of our MTJ-SNG as an architectural construct for ANNs in the SC domain. We develop an optimization algorithm, that approximates the synaptic weights in a 1-layer NN, for achieving energy-efficiency by sacrificing little accuracy.
\item This algorithm is then extended to a multi-layer NN by heuristically breaking down the entire problem into separate problems for each layer and solving them optimally.
\end{itemize}
\IEEEpubidadjcol
%The rest of the paper is organized as follows. Section \ref{prelims} describes the basics of Neural Networks and Stochastic Computing. Section \ref{MTJ} discusses the characteristics of MTJs with regard to switching time and energy, and introduces a low-power MTJ-SNG. This is then used in section \ref{mtj_based_nn} for an energy-efficient NN implementation where an optimization algorithm is proposed to trade accuracy off for considerable power savings. Finally section \ref{results} analyzes the effect of this algorithm using several benchmarks and section \ref{conclusion} concludes the work.
\section{Preliminaries}
\label{prelims}
This section explains the basics of Neural Networks relevant to this paper, their structure and functioning, and mentions prior work similar to ours.

\subsection{Neural Network Architecture}
\begin{figure}[t]
%\centering
\begin{minipage}{.13\textwidth}
\subfloat[]
 {
\includegraphics[height = 1.3cm, width = 1.00\textwidth, trim = 3.2cm 21cm 10.4cm 2.1cm , clip]{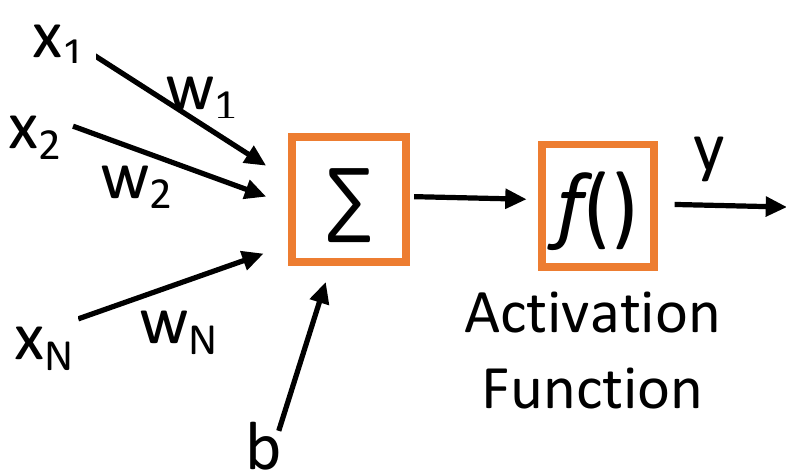}
\label{neuron_schematic}
}  \par
\subfloat[]
 {
\includegraphics[height = 1.5cm, width = 1.00\textwidth, trim = 3.5cm 7cm 15cm 6.5cm , clip]{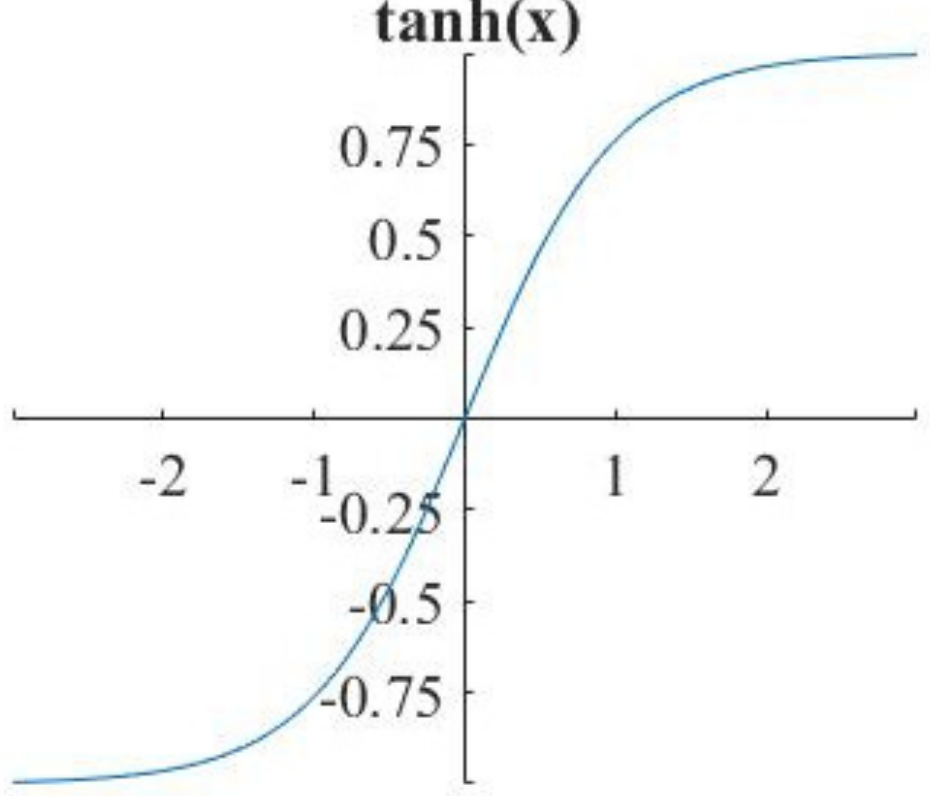}
\label{tanh_plot}
}
\end{minipage}
\hfill
\begin{minipage}{.33\textwidth}
\subfloat[]
{
\includegraphics[height = 3.5cm, width = 0.97\textwidth, trim = 1.0cm 13.0cm 1.4cm 2.5cm , clip ]{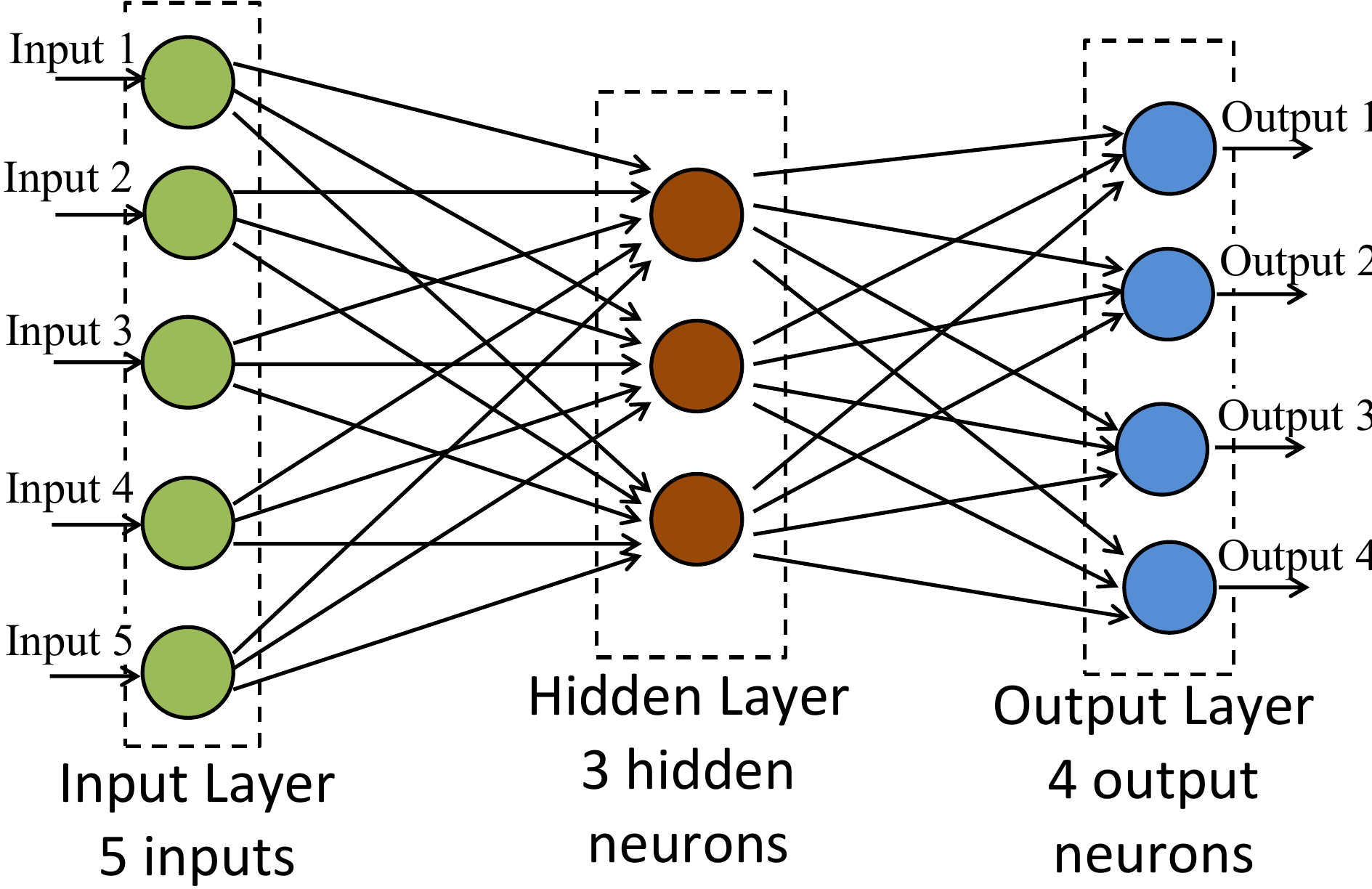}
\label{neural_network}
}
\end{minipage}
\caption{\small (a) A neuron. (b)The $tanh$ function. (c) Schematic of an MLP with one hidden layer.}
\label{nn}
\end{figure}

The fundamental units of an NN are \emph{neurons}, which represent non-linear, bounded functions, and \emph{synapses}, which are 
interconnections between neurons. Each neuron performs a weighted sum of its inputs, which in turn is fed to a non-linear activation 
function to squash the output to a finite range \cite{samarasinghe2016neural}. The output of a neuron, called the \textit{activation level}, 
can be expressed as
\begin{equation}
\label{neuron}
y = f\left(\sum\limits_{i=1}^{N} w_{i}x_{i} + b\right)
\end{equation}
where $N$ is the no. of inputs to the neuron, $w_{i}$ is the synaptic weight of the connection from the $i^{th}$ input $x_{i}$, $b$ is 
a bias, and $f()$ is an activation function (such as $tanh$ or sigmoid). Fig. \ref{nn}\subref{neuron_schematic} depicts the operations performed 
by a neuron and \ref{nn}\subref{tanh_plot}, the behaviour of the $tanh$ function.

\emph{Feedforward networks} are the most elementary Neural Networks, in which information flows only in one direction from the 
input to the output, represented by an acyclic graph. The simplest feedforward network, called a Perceptron, contains just the input and output layers. 
More popular and useful are the Multi-layer Perceptrons (MLPs) which have one or more layers of neurons, called hidden layers, between 
the inputs and the outputs (fig. \ref{nn}\subref{neural_network}).

%The ability of an NN to learn is what makes it useful. Prior to using in applications such as function approximation and classification, an 
%NN has to be trained using several examples, which are pairs of inputs and their corresponding outputs. The weights are initialized to 
%random values and then adjusted as the network is trained to perform a certain task. 
%Weight updates can occur either after each training 
%example is scanned (online learning) or after all of them are scanned (batch training). One single pass/iteration through the entire training 
%dataset is called an \emph{epoch}. This is called Supervised learning, as opposed to Unsupervised Learning, where correct outputs are not 
%known prior to training.
The most popular technique of training an NN is the \emph{error back-propagation} method, which relates the error 
or cost function with the weights of all the layers. This kind of a \textquotedblleft backward calculation\textquotedblright \  is used to compute the gradient of the error 
function that is then used to update the weights in the direction in which error goes down the steepest \cite{samarasinghe2016neural}. This is known as gradient descent 
or the delta rule and is given as
\begin{equation}
\label{gradient_descent}
\Delta w_{i} = -\eta \frac {\partial E}{\partial w_{i}}
\end{equation}
where $ \eta$ is known as the learning rate and $E$ is the error function.

\vspace{-0.5mm}
\subsection{Stochastic Computing}
\label{intro to SC}
In contrast to conventional arithmetic computing, SC uses bit streams to represent numbers, typically denoted by the probability of \textquoteleft1\textquoteright s in the stream. An SN with value $p \in [0,1]$ is represented as a sequence of bits, 
such that if there are $n$ bits in the sequence, out of which $k$ are \textquoteleft1\textquoteright, then $p = \frac{k}{n}$ \cite{alaghi2013survey}. This is known as the \emph{unipolar} 
format. In the \emph{bipolar} format, $p \in [-1,1]$, and the same bit sequence would now have the value $p = \frac{2k-n}{n}$. For 
example, the bit stream 0100101000 would be interpreted as 0.3 in the unipolar format and $-0.4$ in the bipolar format. 

\begin{figure}[b]
%\centering
\subfloat[]
{
\includegraphics[ height = 1.1cm, width = 0.06\textwidth, trim = 3.5cm 22.0cm 15.6cm 2.6cm , clip]{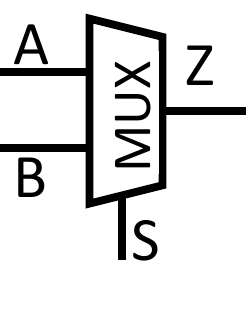}
\label{scaled_addition}
}
\hfill
\subfloat[$0.75 + 0.5 = 1.25$]
{
\includegraphics[  width = 0.18\textwidth, trim = 3.2cm 23.5cm 10.0cm 2.3cm , clip]{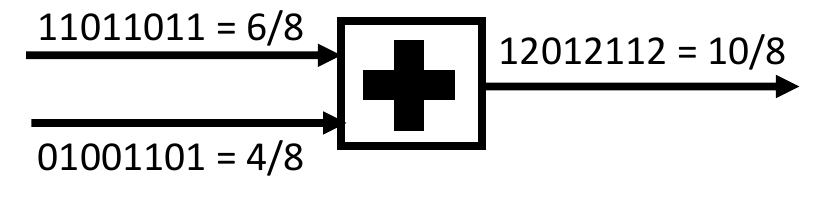}
\label{ISC_add}
}
\hfill
\subfloat[$1.25 \times 1.5 = 1.875$]
{
\includegraphics[ width = 0.18\textwidth, trim = 3.2cm 23.5cm 10.0cm 2.3cm , clip]{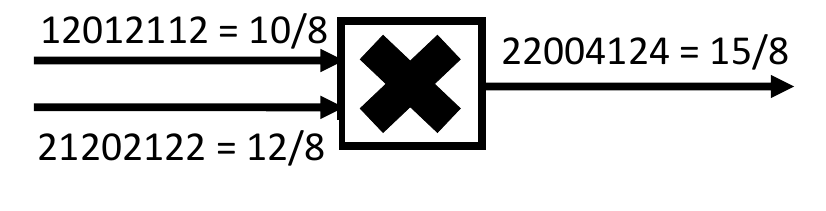}
\label{ISC_mult}
}
\caption{\small (a) Scaled addition in SC, (b) Integral SC (ISC) representation $(m=2)$, and (c) Multiplication in ISC $(m_1 = m_2 = 2)$}
\label{ISC_ops}
\end{figure}

In SC, multiplication is performed by an AND gate in the unipolar format \cite{alaghi2013survey}. Thus, given 2 stochastic streams X and Y, their product is AND(X,Y). 
In the bipolar format, it is given as XNOR(X,Y). However, it is not possible to perform a precise addition in the SC domain as the sum of 2 
SNs might very well lie beyond the range. Only a scaled addition is possible which is achieved through a 2:1 Mux whose Select input is the
scaling factor and is also an SN. The scaled additon of A and B, with scaling factor S, would give Z = A.S + B.(1-S) as in fig. \ref{ISC_ops}\subref{scaled_addition}. With $S = 0.5$, one can 
get $\frac{A+B}{2}$, albeit with a loss of precision.
However, most implementations of NNs involve the sum of a large number of numbers and a loss of precision would only result in  severe 
errors at its outputs.

To overcome this issue, Ardakani et al. \cite{ardakani2016vlsi} introduce the concept of Integral Stochastic Computing (ISC) which allows us to represent 
numbers beyond the range of conventional SC. In the unipolar format, a real number $s \in [0,m]$ can be expressed as the sum of $m$
numbers $s_1, s_2, ... s_m \in [0,1]$. Each of the $s_{i}$\footnotesize s \normalsize  can be represented as stochastic streams and $s$ can be obtained as 
the bit-wise summation of these $m$ streams as illustrated by an example in fig. \ref{ISC_ops}\subref{ISC_add}.
%For eg., 1.25 can be expressed as 0.75 + 0.5 which have 8-bit stochastic representations (say) 11011011 and 
%01001101 respectively. Now, the integral stoschastic stream of 1.25 can be obtained by a bit-wise summation of these, which is 12012112, also represented using 
%2 streams. 
In general, a number $s \in [0,m]$, when represented as the sum of $m$ SNs, would require $\lceil log_{2}m \rceil + 1$ streams (similar to a binary representation). This concept extends similarly to the bipolar format as well \cite{ardakani2016vlsi}.

Multiplication and addition in ISC are performed using binary radix multipliers and adders respectively. Given 2 real numbers $s_1 \in [0,m_1]$ 
and $s_2 \in [0,m_2]$, their product and sum would have $\lceil log_{2}(m_1m_2) \rceil + 1$ and  $\lceil log_{2}(m_1 + m_2) \rceil + 1$ bits respectively  in the ISC domain. Fig. \ref{ISC_ops}\subref{ISC_mult} gives an example. It is also possible to design a good approximation to non-linear functions, such as $tanh$, in ISC using a Finite State Machine \cite{ardakani2016vlsi}.

\subsection{Related Work}
Several research efforts have been made both towards the efficient implementation of deep neural networks through approximations (as in \cite{mrazek2016design}, \cite{zhang2015approxann} and \cite{venkataramani2014axnn})
and towards the realization of NN hardware with non-conventional methods of computation.
%In \cite{mrazek2016design}, Mrazek et al. provide a methodology for designing a power-efficient NN with a uniform structure using approximate multipliers. 
%The key constraints governing the approximation were determined through an error resilience analysis.
%%, with the acceptable error types being specified by a gate level description of the accurate circuit. 
%The algorithm basically performs a design space exploration with the search being guided by an error metric.
%Zhang et al. \cite{zhang2015approxann} propose an approximate computing framework that considers approximating not only the computations but also the 
%memory accesses. They developed an optimization procedure that assesses the criticality of each neuron in terms of its impact on output quality and energy consumption.
%% and that jointly considers error-tolerance capability and energy consumption.
%In \cite{venkataramani2014axnn}, Venkataramani et al. use backpropagation itself to analyze the neuron criticality as the process of training is of an error-healing 
%nature. They use precision scaling to design approximate "low-impact" neurons and the weights connected to them, and design a quality-configurable Neuromorphic Processor Engine that provides a programmable hardware for implementing the approximate
%NNs.
%In the other direction, 
Tarkov \cite{tarkov2015mapping} proposes using a memristor as a device that stores synaptic weights, thereby obviating the 
need for a large amount of memory, and develops an algorithm for mapping a weight matrix onto a memristor crossbar. Hu at al. \cite{hu2016dot} 
extend this idea by developing a Dot-Product Engine (DPE) for matrix-vector multiplications by taking into account the device and circuit 
issues. 
%The speed-accuracy product of the DPE was found to be significantly higher than that of a custom digital ASIC.
In \cite{ardakani2016vlsi} Ardakani et al. design an efficient 
implementation of an NN in the ISC domain. They achieve significant reduction in power consumption at the same rate of misclassification when compared to 
CMOS implementation. Kim et al. \cite{kim2016dynamic} combine the ideas of SC in DNN and energy-accuracy trade-off by removing near-zero weights 
during the training phase (and later retraining the network), combining the addition and squashing operations and incorporating progressive 
precision in the SC bit streams. Venkatesan et al. \cite{venkatesan2015spintastic} proposed a spintronic-based Stochastic Logic which used the random switching 
characteristics of a nanomagnet to generate random numbers and MTJs to store them in binary form.

\section{MTJ-based Stochastic Computing}
\label{MTJ}
In this section we shall describe the probabilistic switching characteristics of an MTJ and exploit the properties of Stochatic Numbers to 
design a low-power optimized MTJ-based SNG and compare it to its non-optimized version. This MTJ-SNG would be the underlying source 
of approximations in our energy-efficient NN implementation.
% Citations here are from the previous paper
\subsection{Characteristics of Magnetic Tunnel Junctions}
MTJ is the most popular spintronic device being considered for NVM technologies \cite{kim2015technology}. 
%It consists primarily of 3 
%layers - two ferromagnetic layers made of CoFeB, and an oxide (typically MgO) layer sandwiched between them acting like a tunnel barrier. 
An MTJ can exist in one of 2 states depending on the relative magnetizations of its free and fixed layers \textendash  \, Parallel (P, logic 0) or Anti-Parallel
(AP, logic 1). MTJs exhibit spin-torque transfer effects \textendash \, spin polarized 
current, when passed through an MTJ, can switch the magnetization of its free layer. Depending on the switching pulse width, MTJs exhibit 
3 switching modes \textendash  \,  Precessional ($<$ 3 ns), Dynamic Reversal (3 to 10 ns) 
and Thermal Activation ($>$10ns) \cite{diao2007spin}. Since we desire a high-speed SNG, we operate in the Precessional mode (high currents leading to switching times of the order of few \textit{ns}), where the probability density 
function of the switching time is given as\footnote{$H_{K}$ is the shape anisotropy field, $M_{s}$ is the saturation magnetization, $V$ is 
the volume of the free layer, $k_{B}$ is the Boltzmann constant, $T$ is the temperature, $J$ is the current density, $J_{c0}$ is the critical 
current density, $\eta$ is the spin transfer efficiency, $t_{F}$ is the thickness of the free layer and $t_{p}$ is the pulse duration.}
\begin{equation}
\label{switching_pdf}
P(t_{p}) \propto e^{-\Delta sin^{2}\phi}(J - J_{c0})sin^{2}\phi
\end{equation}

\noindent with $\Delta = \frac{H_{K}M_{s}V}{2k_{B}T}$ and $\phi = \frac{\pi}{2}e^{-\frac{\eta \mu_{B}}{eM_{s}t_{F}} (J-J_{c0})t_{p}}$

\vspace{2mm}
We have simulated the behavior of an MTJ with in-plane magnetic anisotropy using an MTJ Spice Model\footnotemark  \cite{kim2015technology}. The values of 
$J_{c0}$ obtained were $ 7.55 MA/cm^2$ for P$\rightarrow$AP switching and $4.10 MA/cm^2$ for AP$\rightarrow$P switching and of 
$\Delta$ was 47.5.% (sufficient retention for our purpose).

\footnotetext{The parameters used were - cell dimension $20 \times 58 nm^2$, $t_F = 2.5  nm$, $M_s = 1222  emu/cc$, 
$\alpha$ (damping constant) $= 6.82 X 10^{-3}$, $\eta= 0.85$, RA product $= 5 \Omega \mu m^2$, $T = 300K$.}

Given a pulse of width $T_p$, the probability that switching takes place within $T_p$ is \vspace{-1mm}
\begin{equation}
\label{switching_cdf}
P_{sw}(T_p) = \int_{0}^{T_p} P(t) dt
\end{equation}
\noindent where $P_{sw}(t)$ is the switching probability (the cumulative distribution function), and the expected time at which switching 
takes place (given it does) with pulse width $T_p$, is expressed as
\begin{equation}
%\label{switching_expected_time}
E(t_{sw}) = \int_{0}^{T_p} t P(t) dt
\end{equation}
Let $I_{AP}$ and $I_P$ denote the currents in the AP and P state respectively. The expected energy consumed in such a scenario, for 
AP$\rightarrow$P switching, is
\begin{equation}
%\label{switching_cdf}
E^{AP\rightarrow P}_{sw} =  V\left(I_{AP}E(t_{sw}) + I_{P}(T_{p} -  E(t_{sw}))\right) 
\end{equation}
\noindent whereas the energy spent in the case where switching does not take place is \vspace{-2mm}
\begin{equation}
%\label{switching_cdf}
E^{AP\rightarrow P}_{nsw} =  VI_{AP}T_p
\end{equation}
\noindent Thus the overall expected energy consumed is given as
\begin{equation}
%\label{switching_cdf}
E = P_{sw}(T_p)E^{AP\rightarrow P}_{sw} + (1- P_{sw}(T_p))E^{AP\rightarrow P}_{nsw}
\end{equation}

\subsection{MTJ as a Stochastic Number Generator}

\begin{figure}[t]
\centering
\subfloat[Probability v/s pulse width]
 {
\includegraphics[height = 3.2cm, width = .21\textwidth, trim = 0cm 0cm 1cm 2.3cm , clip]{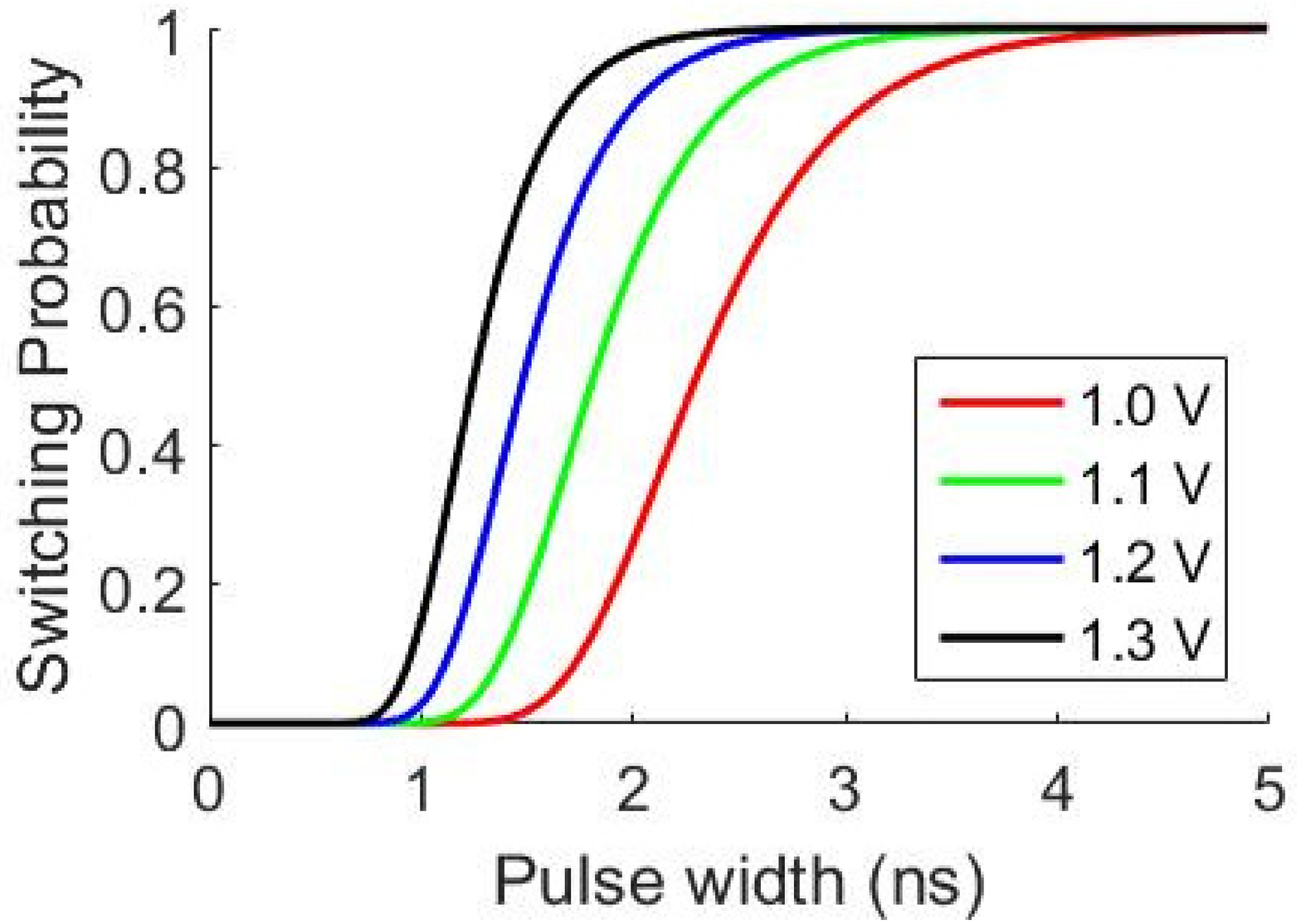}
\label{MTJ_probability}}
\hfill
\subfloat[Energy v/s probability, $V_{bias} = 1.2V$]
 {\includegraphics[width = .25\textwidth,  trim = 2.5cm 21.5cm 14cm 2.7cm , clip]{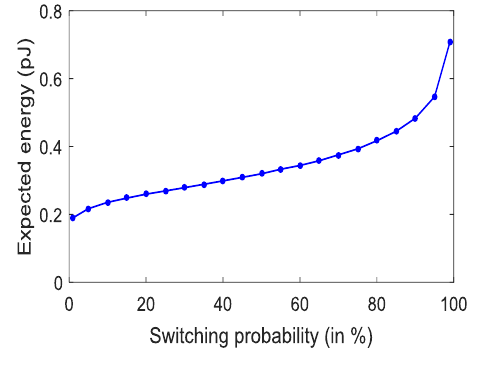}
\label{MTJ_energy}
}

\caption{\small MTJ switching characteristics for $AP\rightarrow P$ transition. }
\label{MTJ switching}
\end{figure}

An MTJ can be used as an SNG by exploiting the probabilistic nature of its switching. Given a voltage pulse, the probability of switching 
can be decided by controlling the pulse width. The probabilities for AP$\rightarrow$P switching, for different voltage bias, are shown in 
fig \ref{MTJ switching}\subref{MTJ_probability}. For each bit generated by the MTJ representing a stochastic number $p \in [0,1]$, one would typically do the following iteratively:

\begin{itemize}
\item[i.] Reset to ‘0’ with 100\%  probability (not required if state didn't change in the previous iteration)
\item[ii.] Write ‘1’ with probability $p$, and
\item[iii.]  Read the value stored in the MTJ (which would be ‘1’ with probability $p$ and ‘0’  with probability $1-p$).
\end{itemize}

\noindent Repeating this procedure $n$ times would give us a sequence of $n$ bits, out of which $p.n$ are expected 
to be 1, thereby representing the SN $p$. However, the expected energy required for switching P$\rightarrow$AP (logic 0$\rightarrow$1) 
with 99.9\% probability is $0.46pJ$; whereas that for AP$\rightarrow$P (logic 1$\rightarrow$0) is $0.93pJ$ with 
1.2V. We thus choose the AP state (logic 1) to be the reset state, and switch to the P state (logic 0) with some probability (because 
resetting P$\rightarrow$AP would require lesser energy than AP$\rightarrow$P). This means that switching AP$\rightarrow$P 
with probability $x$ will generate bit streams where the probability of finding \textquoteleft 1\textquoteright s is $1-x$. Hence, to represent the stochastic number $p$, we 
shall write 0 with probability $1-p$.

The expected energy consumption was found to be minimum for a voltage bias of about 1.2V. We thus use this for writing to the MTJ. 
Plotted in fig. \ref{MTJ switching}\subref{MTJ_energy} is the trade-off between energy and switching probability. Resetting the MTJ requires a pulse width of 
$4.33ns$, and switching to \textquoteleft0\textquoteright \ with 99.9\% probability requires $3.40ns$. Reading the value stored in the MTJ using a sense 
amplifier can be done with a bias of $-0.1V$ for $2 ns$. Thus, the total time necessary for generating one bit of the SN is (a maximum of) 
$9.73 ns$.

\subsection{Proposed Biased MTJ-SNG}
\label{BMS}

\begin{figure}[t]
\centering
\subfloat[]
 {
\includegraphics[width = .18\textwidth, trim = 2.3cm 19cm 8.6cm 1.5cm , clip]{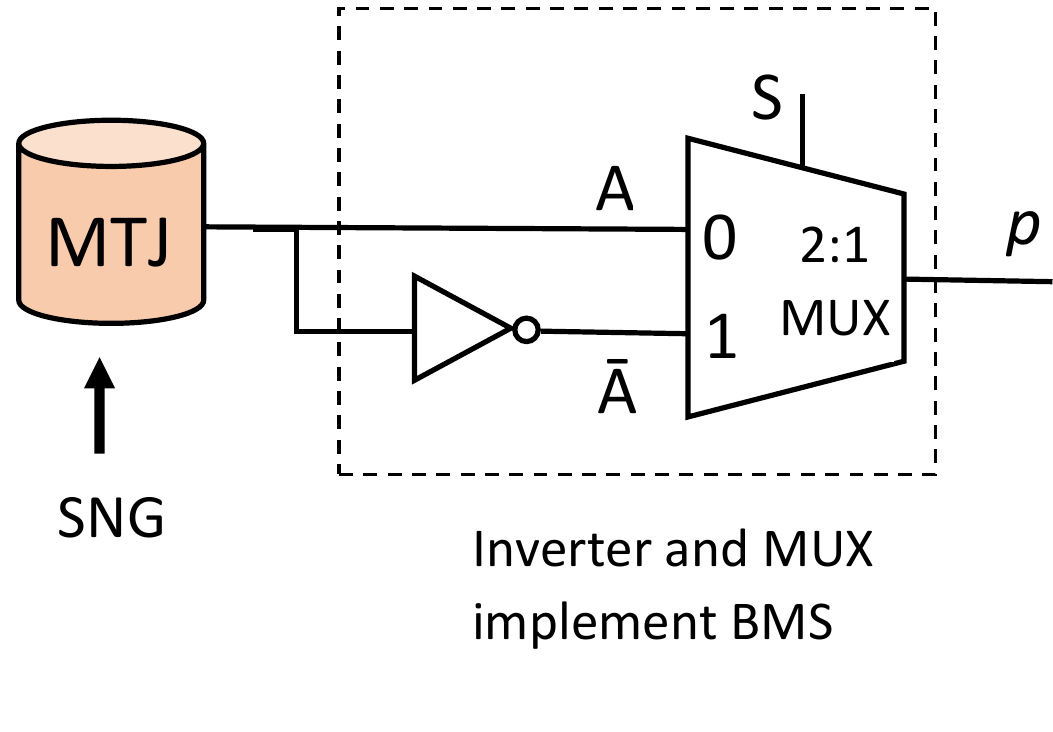}
\label{bms_circuit}
}
\subfloat[]
 {\includegraphics[height = 2.8cm, width = .28\textwidth, trim = 0cm 2cm 1.5cm 3cm , clip ]{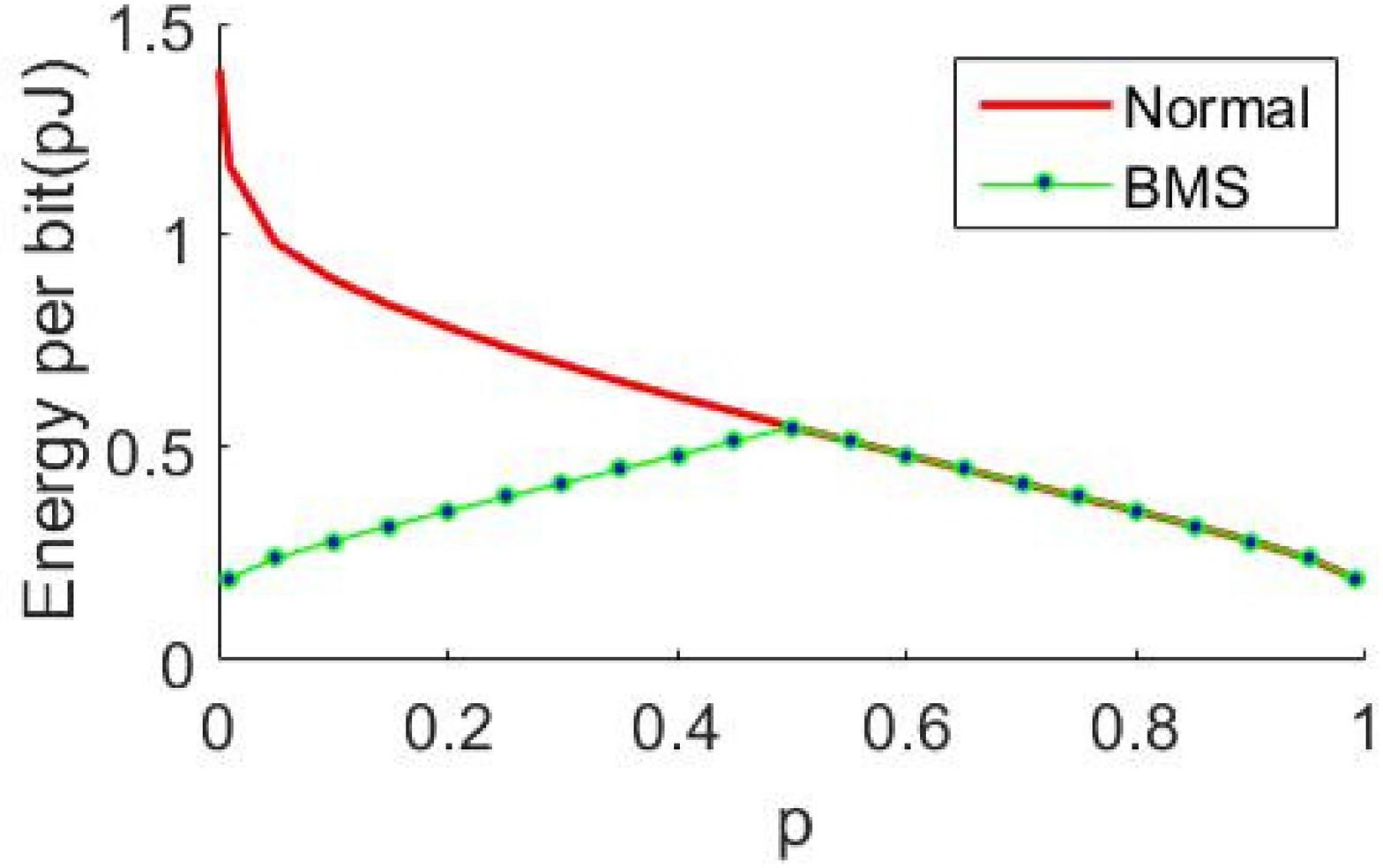}
\label{bms_energy}
}
\caption{\small (a) Circuit of the BMS. (b) Energy per bit v/s Value of SN $p$ }
\label{bms}
\end{figure}

We make a slight modification to the overall procedure of generating the bits of the SN. As seen earlier, generating an SN with value $p$ 
using our MTJ implies that the probability of switching from AP$\rightarrow$P (which is the same as writing ‘0’) has to 
be $1-p$.  If $p$ is closer to 0, then $1-p$ is closer to 1; which implies more time, and hence more energy, has to be spent in writing \textquoteleft0\textquoteright \ to the MTJ, as compared to the case 
where we had to generate an SN with value $1-p$. To prevent this characteristic from making the SNG energy-intensive, we choose to
generate $1-p$ whenever $p < 0.5$ (but generate $p$ if $p \geq 0.5$). In other words, whenever $p < 0.5$, instead of switching AP$\rightarrow$P with probability $(1-p)$ (which is $\geq 
0.5)$, we switch with probability $p$. Now all we would need to do is to invert the bits output from this Biased MTJ-SNG 
(\textbf{BMS}, the name being derived from the biased nature of the data produced by the MTJ-SNG) so that we get back the SN $p$. Therefore, we generate either $p$ or $1-p$, whichever is larger, and use a 2:1 multiplexer 
to choose between the generated SN and its inverse as shown in fig. \ref{bms}\subref{bms_circuit}. The inputs to the Select pin of the multiplexer can be derived from the 
most significant bit of the binary number that is being converted to a stochastic number \cite{alaghi2013survey}. As an example, if $p = 0.7$, the MTJ-SNG 
will generate $p$ itself and S will be 0 to output $A = 0.7$. On the other hand, if $p = 0.3$, the MTJ will generate $(1-p) (= 0.7)$ and 
S will be 1 to output $\bar{A} = 0.3$.

The energy required to generate one bit from the MTJ-SNG is plotted in fig. \ref{bms}\subref{bms_energy} as a function of $p$ for both the cases - without any modification (Normal), and with BMS in place (including the overheads due to the mux and inverter). The symmetry of the plot 
with BMS (dotted line) comes from generating the larger of $p$ and $1-p$. Table \ref{normal_vs_bms} compares the 2 MTJ-SNGs. Since the BMS requires us to generate SNs only 
greater than or equal to 0.5, the maximum write duration 
reduces from $3.40 ns$ to $1.49 ns$ (the latter corresponds to the pulse width giving 50\% switching probability), thereby decreasing the total time. The average energy and power have been calculated considering a uniform distribution of $p$ over the range [0, 1]; BMS brings about a reduction by 42\% and 23\% respectively (without introducing any approximation or error in the SN being generated.)

\small
\begin{table}[h]
\caption{\footnotesize Comparison of Normal and Biased MTJ-SNG}
\label{normal_vs_bms}
\renewcommand{\arraystretch}{1.1}
\centering
\begin{tabular}{|c|c|c|c|}
	\hline
	MTJ-SNG & Time$(ns)$ & Avg. Energy $(pJ)$ & Avg. Power $(\mu W)$ \\
	\hline
	Normal & 9.73 & 0.59 & 80 \\
	\hline
	BMS & 7.82 & 0.34 & 62 \\
	\hline
\end{tabular}
\end{table}
\normalsize

\section{Energy Efficient MTJ-based NN Implementation}
\label{mtj_based_nn}
Stochastic circuits have gained popularity in low-cost implementation of NNs \cite{ardakani2016vlsi} \cite{kim2016dynamic}. We propose using MTJs as a hardware 
component for realizing NNs in the SC domain by exploiting their probabilistic switching nature to generate SNs representing inputs and synaptic weights. 
The error-resilient nature of NN applications motivate us to approximate the weights, effectively designing approximate multipliers, and thereby gaining energy efficiency. 
In this section, we develop an algorithm that, given a trained network, the training dataset and an error tolerance, adjusts the weights in the 
best possible way in the solution space, while remaining within the error constraint at all times.

\subsection{NN implementation in the SC/ISC domain}

\begin{figure}[b]
\centering
\subfloat[]
{
\includegraphics[width = 0.30\textwidth, trim = 3.6cm 18.8cm 3.2cm 2.6cm , clip]{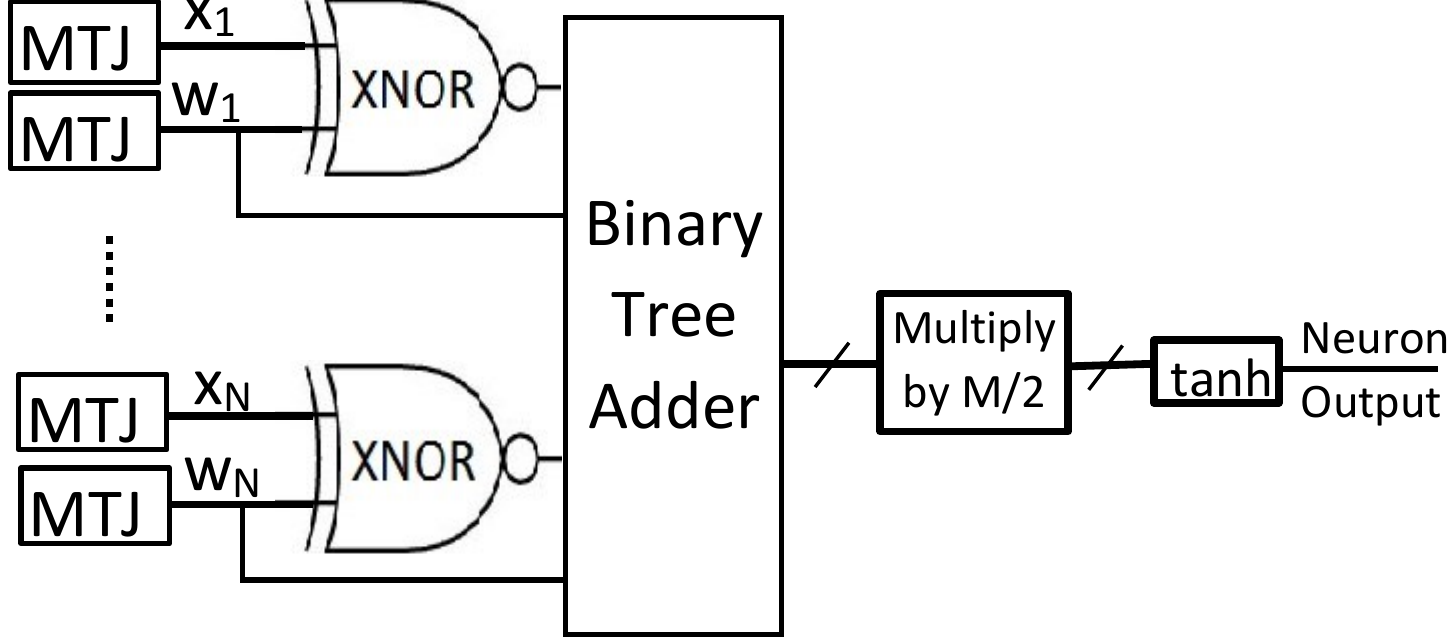}
\label{neuron_ISC}
}
\hfill
\subfloat[]
{
\includegraphics[width = 0.15\textwidth, trim = 2.6cm 20.3cm 12.0cm 2.0cm, clip]{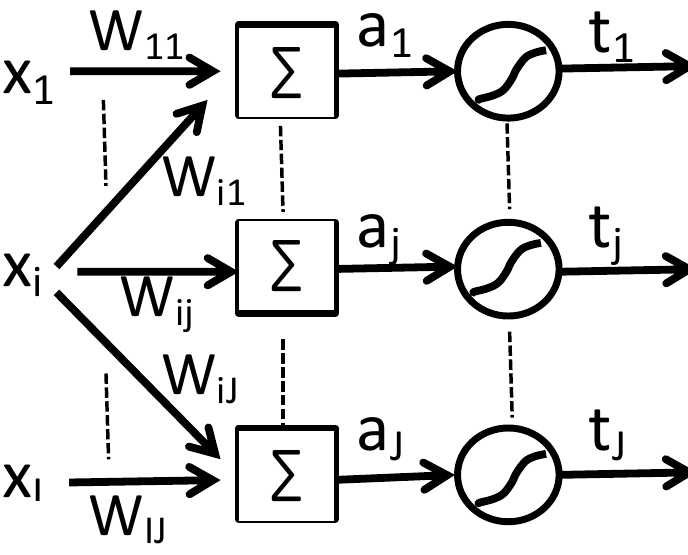}
\label{1_layer_schematic}
}
\caption{\small (a) Neuron implementation in ISC (b) Schematic of 1-layer NN}
\label{ISC NN}
\end{figure}

Here we describe how the operations of a neuron would be performed in the ISC domain (refer to section \ref{intro to SC}). We know that the activation level of a neuron is given as
\begin{equation}
\label{activation_level}
t = f(a) = f\left(\sum\limits_{i=1}^{N} \tilde{w}_{i} \tilde{x}_i\right)
\end{equation}
where $f$ is the activation function operating on $a$, the weighted sum of inputs. Several types activation functions can be used in an NN. We go for the $tanh$ function because it is non-linear and has a bipolar output .
%\begin{itemize}
%\item It limits the output to finite range 
%\item It is continuous and differentiable, enabling us to use the delta rule in error back-propagation
%\item It mimics the behaviour of biological neurons, and so is the most popular one.
%\end{itemize}
In eqn. (\ref{activation_level}), the $\tilde{x}_i$\small s \normalsize (inputs) are assumed to be in the range $[0,1]$ (and are typically so) and $\tilde{w}_{i}$\small s \normalsize (weights) are generally in 
$[-M,M]$ with $M > 1$. The latter can be 
represented in the ISC domain with $\lceil \log_2M \rceil + 1$ stochastic streams; however, this would need those many SNGs, leading to high 
energy consumption. Therefore, we have to scale them down to the range $[0,1]$ or $[-1,1]$ to be able to use only 1 stream. Since the ISC 
implementation of the $tanh$ function using FSM is in bipolar format, we go for the interval $[-1,1]$. Further, it is necessary to have a single 
format throughout, thereby requiring us to scale down the inputs to the range $[-1,1]$. So the weighted sum would now be written as
\begin{equation}
\label{activation_bipolar}
a = \frac{M}{2}\left(\sum\limits_{i=1}^{N} w_{i}x_i + \sum\limits_{i=1}^{N} {w}_{i} \right)
\end{equation}
where  $x_i, w_i \in [-1,1] \ \forall \ i$. Fig. \ref{ISC NN}\subref{neuron_ISC} illustrates the operations of a neuron in the ISC domain, implementing eqns. (\ref{activation_level}) and (\ref{activation_bipolar}). Several such neurons in parallel would form a layer as in fig. \ref{nn}\subref{neural_network}, and multiple layers connected in series would make up the entire network. Note that the output of the $tanh$ is a single stochastic stream in the bipolar format.

\subsection{Problem Formulation}

As can be seen from fig. \ref{bms}\subref{bms_energy}, the generation of SNs (from the proposed BMS) that are closer to 0 or 1 require less energy as compared to those that 
are closer to 0.5. In the bipolar format of SC, this would imply low energy requirement for numbers closer to 1 or $-1$ than to 0. This property of the BMS forms the basis of achieving energy-efficiency through approximations that tend to shift the weights \textquotedblleft farther from\textquotedblright \  0 towards 1 or $-1$, whichever is closer. We therefore aim to bring the weights of the network as close to 1 or $-1$ as possible while ensuring that output errors are within a tolerance level for all the training inputs. We investigate both single-layer and multiple-layer NNs.

%We therefore use a first-order approximation of the error in the neuron output as shown below (the second-order approximation is again non-convex).
%Given a continuous and differentiable function $f(x)$, we know from Taylor series approximation, that
%\begin{equation}
%\label{Taylor}
%f(x + dx) = f(x) + f'(x)dx + \frac{f''(x)}{2}dx + ...
%\end{equation} 
%
%\noindent A first order approximation would therefore yield
%\begin{equation}
%\label{FOA}
%f(x + dx) - f(x) = f'(x)dx 
%\end{equation} 
%
%\noindent The derivative of the hyperbolic tangent is given as
%\begin{equation}
%\label{derivative}
%\frac{d}{dx}tanh(x) = \frac{4}{(e^x + e^{-x})^2}
%\end{equation}
%
%\noindent which shall be used in our formulation.

\subsection{Optimizing a 1-layer NN}
For a single layer network, we illustrate how to formulate the approximation as a convex optimization problem. Convexity of the feasible region of such a problem implies that any local minimum in that region is also the global minimum, ensuring that the optimum value of the objective function is always achieved. Further, non-convex optimization problems are more complicated to solve.

The objective of our formulation is to minimize the separation of the weights from 1 or $-1$ (whichever is closer). Since the weights are independent of each other, the objective function can be expressed as the sum of the \textquotedblleft distance\textquotedblright \ of the weights from 1 or $-1$. One way of specifying an error tolerance at the output layer is to measure the deviation of the output neurons from their actual values (the values obtained from 
the trained network) and restrict all of them to within some threshold. Such a constraint should be also be applicable for all input vectors used in the optimization. 

However, the $tanh$ function (which provides the neuron output) is not only non-linear but 
also non-convex. Thus, neither neuron activation levels nor the errors in them can be directly incorporated in the convex formulation. 
But the input to this activation function is affine (hence convex) because 
it is a weighted sum of inputs. 
We therefore need to translate the output errors to errors in inputs of $tanh$. Given a limit to the deviation in neuron output, we pre-compute the upper and lower limits of the weighted sum input using the $tanh^{-1}$ function and force it to remain within these limits.

Fig. \ref{ISC NN}\subref{1_layer_schematic} illustrates a 1-layer network having $I$ inputs and $J$ outputs and table \ref{notation_1layer} lists the notations. In addition, the presence of \ $\hat{}$ (hat)  symbol indicates that 
the quantity is the original value obtained from the trained network, and hence is a constant in the problem; whereas its absence denotes 
a variable. 
%From  (\ref{FOA}) and (\ref{derivative}), the first-order approximation of the change in neuron output would be given as
%\begin{equation}
%\label{FOA_tanh}
%t^{r}_j - (\hat{t}^{r})_j =\frac{4}{\left(e^{\hat{a}^{r}_j} + e^{-\hat{a}^{r}_j}\right)^2}\left({a}^{r}_j - \hat{a}^{r}_j\right)
%\end{equation}
%where $t^{r}_j$ denotes the output at the $j^{th}$ neuron for the $r^{th}$ training data sample and ${a}^{r}_j$ is the corresponding weighted sum input as illustrated in fig. \ref{1_layer_schematic}.

%%Notations
\begin{table}[h]
%% increase table row spacing, adjust to taste
%\renewcommand{\arraystretch}{1.3}
% if using array.sty, it might be a good idea to tweak the value of
% \extrarowheight as needed to properly center the text within the cells
\caption{\footnotesize Notations for problem formulation of 1-layer NN}
\label{notation_1layer}
\centering
%% Some packages, such as MDW tools, offer better commands for making tables
%% than the plain LaTeX2e tabular which is used here.
\setlength{\tabcolsep}{1mm}
\renewcommand{\arraystretch}{1.1}
\begin{tabular}{|c|c|c|c|}

\hline
Name & Meaning & Type & Dimension\\
\hline
$W$ & The output layer weight matix  & Matrix & $I \times J$\\
\hline
$\hat{x}^r$ & The $r^{th}$ training sample (input vector)& Vector & $I$\\
\hline
$M$ & The scaling factor for $W$ & Scalar & $1$\\
\hline
$a^r$ & The $r^{th}$ weighted sums (output layer) & Vector & $J$\\
\hline
$t^r$ & The $r^{th}$ activation levels (output layer) & Vector & $J$\\
\hline
\end{tabular}
\end{table}
% Notations end here

\begin{algorithm}[b]
\footnotesize
\caption*{\textbf{\small Weight Approximation for a single-layer NN}}
\label{ConvexOpt}
\begin{algorithmic}[1]
\Procedure {OptimWeights}{$I$,$J$,$\hat{W}$, $\hat{x}$, $\hat{t}$, $M$,$\phi$}

\smallskip
\State The constraint on the neuron outputs are $\left |{t}^{r}_j - \hat{t}^{r}_j\right | \leq \phi$. Compute the \Statex \quad \ \ upper and lower limits of all weighted sums as \label{upper and lower limits}
\Statex \quad \ \ $u^{r}_{j} = tanh^{-1}\left(\hat{t}^{r}_j + \phi \right) $ and
\Statex \quad \ \ $v^{r}_{j} = tanh^{-1}\left(\hat{t}^{r}_j - \phi \right) $ respectively
\hfill $\forall$ \ $ r = 1 \ldots D$,  $ j=1\ldots J$  

%\smallskip
%\State Compute the upper limits as $u^{r}_{j} = tanh^{-1}\left(\hat{t}^{r}_j + \phi \right) $ \label{upper limit}
%
%\State Compute the lower limits as $v^{r}_{j} = tanh^{-1}\left(\hat{t}^{r}_j - \phi \right) $ \label{lower limit}

\smallskip
\State Solve the optimization problem: \label{objective}
    \Statex \quad \ \  $\underset{W}{\textbf{minimize}} \ W_{sod} = \sum\limits_{i=1}^{I} \sum\limits_{j=1}^{J} {W^{\prime}_{ij}} $  \label{weight_sum}
    \Statex \quad \ \ \textbf{subject to} the following constraints (lines \ref{Wconstraint} to \ref{restrict weighted sums}): 

%\State
%$ \begin{IEEEeqnarraybox}[][c]{l?s}
%\IEEEstrut
%   0 \leq W_{ij} \leq 1  &  if  $\hat{W}_{ij} \geq 0 $ \\
%-1 \leq W_{ij} \leq 0 & $O.W.  \ \ \ \ \ \  \forall \ i=1 \ldots I,\ j=1 \ldots J$
%\IEEEstrut
%\end{IEEEeqnarraybox} $  \label{Wconstraint}

%\If{$\hat{W}_{ij} \geq 0 $} 
%	\State $0 \leq W_{ij} \leq 1$
%\Else
%	\State $-1 \leq W_{ij} \leq 0$
%\EndIf
\smallskip
\State Restrict the weights to their original range.  \label{Wconstraint}
\smallskip
\Statex \quad \ \ \textbf{if} $\left(\hat{W}_{ij} \geq 0 \right)$ \textbf{then} \ \   $0 \leq W_{ij} \leq 1$
\Statex \qquad \qquad \qquad \qquad \ \textbf{else} $-1 \leq W_{ij} \leq 0 \quad \forall \ i=1 \ldots I,\ j=1 \dots J$

\smallskip
\State See how far the weights are from $1$ or $-1$, whichever is closer
\Statex
\quad \ \ $W^{\prime}_{ij} = \left\{ \,
\begin{IEEEeqnarraybox}[][c]{l?s}
\IEEEstrut
1 + W_{ij}  & $if \ {W}_{ij} \leq 0$, \\
1 - W_{ij} & $otherwise \quad \quad  \forall \ i=1 \ldots I,\ j=1 \dots J$
\IEEEstrut
\end{IEEEeqnarraybox}
\right.$ \label{Wdash}

%\smallskip
%\Statex \quad \ \ It effectively implements ${W}^{\prime}_{ij} = min(1 + W_{ij} , 1 − W_{ij} )$; however this
%\Statex \quad \ expression cannot be directly used as the minimum of affine functions is not convex [15].
%

\smallskip
\State Compute the weighted sum to all neurons for all input vectors
\smallskip
\Statex \ \ \ \ $a^r = \frac{M}{2}(W^{T}\hat{x}^r + W^{T}\textbf{1}) \quad \forall \  r = 1  \ldots D$ \label{activation}

\smallskip
\State Constrain these weighted sums within their upper and lower limits
\smallskip
\Statex \ \ \ \ $v^{r}_j \leq a^{r}_j \leq u^{r}_j $ \label{restrict weighted sums}
\quad $\forall$ \ $ r = 1 \ldots D$,  $ j=1\ldots J$  
%\State $\Delta a^r = a^r - \hat{a}^r \hfill  \ldots \forall \  r = 1  \ldots D$  \label{activation_change}
%\State $\Delta t^{r}_j = f^{\prime} (\hat{a}^{r}_j) \Delta a^{r}_j  \hfill  \ldots \forall \ r = 1  \ldots D,\ j=1 \ldots J$  \label{output_change}
%\State $|\Delta t^{r}_j| \leq \phi$     \hfill   \ldots   $\forall$ \ $ r = 1 \ldots D$,  $ j=1\ldots J$   \label{output_constraint}
%\State $W_{sod} = \sum\limits_{i=1}^{I} \sum\limits_{j=1}^{J} {W^{\prime}_{ij}} $  \label{weight_sum}
\smallskip
\EndProcedure
\State \textbf{return} $t^r = tanh(a^r)$ \quad  $\forall$ \  $ r = 1 \ldots D$ \label{approximated inputs}
\end{algorithmic}
\end{algorithm}
\normalsize

\textbf{The Optimization Procedure: }The procedure for approximating weights in a 1-layer NN is shown below. It takes a trained network 
and an error threshold $\phi$ as inputs, and minimizes the \textquotedblleft sum of distances\textquotedblright \  using $D$ samples of the 
training dataset. Line \ref{upper and lower limits} computes the maximum and minimum values that the weighted sum inputs of the $tanh$ function can take. Here $t^{r}_j$ denotes the output of the $j^{th}$ neuron for the $r^{th}$ training input, and $u^{r}_j$ \& $v^{r}_j$ are the corresponding limits. The objective function (line \ref{weight_sum}) to be minimized is the \underline{s}um \underline{o}f \underline{d}istances of the weights from 1 or $-1$. $ W^{\prime}$ in line \ref{Wdash} stores how far they 
are from 1 or $-1$, whichever is closer. It effectively implements $W^{\prime}_{ij} = min(1 + W_{ij},1 - W_{ij})$; however this expression 
cannot be directly used as the minimum of affine functions is not convex \cite{boyd2004convex}. This is also the reason why we impose a constraint on the range of the weights in line \ref{Wconstraint} (minimum of distance from 1 and $-1$ isn't convex). Line \ref{activation} computes the weighted sum inputs of the $tanh$ function, line \ref{restrict weighted sums} constrains them within the limits obtained in line \ref{upper and lower limits}, and line \ref{approximated inputs} finally returns the approximate neuron outputs. The optimization problem stated above is convex because the objective function and the inequality constraints are convex and the equality constraints are affine \cite{boyd2004convex}.
%After the optimization problem is solved%(line \ref{approx_in}),
%, we calculate the approximated neuron activation levels as $t^r = tanh(a^r)$.
%, which can now be used to check the accuracy of the NN.
%\begin{algorithm}[h]
%\small
%\caption{\small Problem Formulation for 1-layer NN}
%\label{1layer_optimizer}
%\begin{algorithmic}[1]
%\vspace*{-1mm}
%\State $a =$ \Call{OptimWeights}{$I$,$J$,$\hat{W}$, $\hat{x}$, $\hat{a}$, $f'(\hat{a})$, $M$,$\phi$} \label{approx_in}
%\State $t^r = tanh(a^r)$ \hfill   $\ldots \forall$  $ r = 1 \ldots D$ \label{approx_out}
%\vspace*{-1mm}
%\end{algorithmic}
%\end{algorithm}
%\normalsize
%\begin{equation}
%\label{final_output}
%t^r = tanh(a^r) \forall  r = 1 \ldots D
%\end{equation}
%\newtheorem{theorem}{Theorem}
%\begin{theorem}
%The problem formulated above is a convex optimization problem.
%\end{theorem}
%\begin{IEEEproof}
%Need to discuss.
%\end{IEEEproof}
\subsection{2-layer NN}
A similar formulation could have been made for NNs containing more than 1 layer, having the objective of minimizing the \textquotedblleft sum of 
distances\textquotedblright \  of each 
of the weight matrices, with constraints computing the hidden layer(s) outputs and finally restricting the error in the output layer's weighted sums. 
However, the presence of the non-convex activation function in the 
hidden layer(s) would make the problem (as a whole) non-convex. To mitigate this issue, we propose breaking down the problem into separate 
but identical convex problems, each of which optimizes the weights in successive layers of the NN under some error constraints. Thus, in a 2-layer NN having 
$I$ inputs, $L$ hidden neurons and $J$ output neurons, we shall solve 2 problems successively - first for the hidden layer and then for the output layer, 
with error thresholds $\phi_Z$ and $\phi_W$ respectively. Notations used for the 2-layer network, for terms which do not appear in a 
1-layer network, are described in Table \ref{notation_2layer}.
\small
%%Notations
\begin{table}[h]
%% increase table row spacing, adjust to taste
%\renewcommand{\arraystretch}{1.3}
% if using array.sty, it might be a good idea to tweak the value of
% \extrarowheight as needed to properly center the text within the cells
\caption{\footnotesize Notations for problem formulation of 2-layer NN}
\label{notation_2layer}
\centering
%% Some packages, such as MDW tools, offer better commands for making tables
%% than the plain LaTeX2e tabular which is used here.
\setlength{\tabcolsep}{1pt}
\renewcommand{\arraystretch}{1.1}
\begin{tabular}{|c|c|c|c|}
\hline
Name & Meaning & Type & Dimension\\
\hline
$W, Z$ & The output and hidden layer weight matix  & Matrix & $L\times J, I\times L$\\
\hline
%$W$ & The output layer weight matix  & Matrix & L x J\\
%\hline
%$\hat{x}_r$ & The $r^{th}$ input & Vector & I\\
%\hline
%$M_Z$ and & The scaling factor of Z & Scalar & 1\\
%\hline
$M_W$, $M_Z$ & The scaling factor of $W$ and $Z$ & Scalar & $1$\\
\hline
$b^r$ & The $r^{th}$ weighted sums of hidden neurons & Vector & $L$\\
\hline
%$a_r$ & The $r^{th}$ weighted sums of output neurons & Vector & J\\
%\hline
$h^r$ & The $r^{th}$ hidden neuron outputs & Vector & $L$\\
\hline
%$t_r$ & The $r^{th}$ output neuron outputs & Vector & J\\
%\hline
\end{tabular}
\end{table}
% Notations end here
\normalsize
\subsubsection{Estimation of maximum tolerable $\phi_Z$}
In a 2-layer NN, given some value of $\phi_W$, there exists an upper limit to the amount of error that can be tolerated at the outputs of the hidden layer. 
We know that the weighted 
sum input to the $j^{th}$ neuron of the output layer is
\begin{equation}
\label{activation_2layer}
a_j = \sum\limits_{l=1}^{L} W_{lj}h_l 
\end{equation}

\noindent A constraint on the output layer outputs for all of the $D$ inputs is written as
\begin{equation}
\label{constraint_2layerA}
|\Delta t^r_{j}| = |t^r_{j} - \hat{t}^{r}_j| \leq  \phi_W \ \ \ \  \     \forall j=1 \ldots J, r = 1 \dots D
\end{equation}

\noindent We use a first order approximation (from Taylor series expansion) to the errors in the weighted sums and write (\ref{constraint_2layerA}) as
\begin{equation}
\label{constraint_2layerB}
\lvert a^{r}_j - \hat{a}^{r}_j \rvert  \leq   \frac{\phi_W}{f'(\hat{a}^{r}_j)} \ \ \ \  \     \forall j=1 \ldots J, r = 1 \dots D
\end{equation}

\noindent where $f'$, is the first derivative of $tanh$. Because $tanh$ is a monotonically increasing function, $f'$  is always positive. To establish a lower bound, we need to consider the strictest of all constraints, 
which takes us to
\begin{equation}
\label{constraint_2layerB2}
\lvert a_j - \hat{a}_j \rvert  \leq   \min_{r} \left( \frac{\phi_W}{f'(\hat{a}^{r}_j)} \right) = \lambda_j  (say) \ \ \ \ \  \forall j=1 \ldots J
\end{equation}

\noindent Because we are interested in deviations in hidden neuron outputs, using (\ref{activation_2layer}) and then writing $ (h_l - \hat{h}_l) = y_l$, we obtain
\begin{equation}
\label{constraint_2layerC}
\left | \sum\limits_{l=1}^{L} W_{lj}(h_l - \hat{h}_l) \right | = \left | \sum\limits_{l=1}^{L} W_{lj}y_l \right | \leq \lambda_j \ \ \  \forall j=1 \dots J
\end{equation}

\noindent The LHS of (\ref{constraint_2layerC}) represents a hyperplane (for each $j$) in the $L$-dimensional space, with slab constraint 
(\ref{constraint_2layerC}) having L variables and J equations.

\textbf{Case 1:} $L > J$.  The feasible region defined by (\ref{constraint_2layerC}) is unbounded. Thus we can only estimate a lower bound on the 
maximum error tolerable at the hidden layer. The smallest $y$ to violate the $j^{th}$ constraint (say $\tilde{y}_j$) would be orthogonal to the $j^{th}$ 
hyperplane and lying on it. Since each of the $L$ hidden neurons must satisfy the error threshold, we shall consider the largest of the $L$ 
coordinates of $\tilde{y}_j$ for each of the $J$ constraints. The lowest of $J$ such values would provide a lower bound on the maximum accetable 
error. Mathematically, lower bound
\begin{equation}
\label{lower_bound}
\bar{\phi}_Z = \min_{j} \left( \max_{l} \left | \left( \tilde{y}_j \right)_l \right | \right)
\end{equation}

\noindent Because all equations are linear, $\bar{\phi}_Z$ can be obtained quickly through linear programming.

\textbf{Case 2:} $L \leq J$.  The feasible region defined by inequality (\ref{constraint_2layerC}) is bounded. We can find a lower bound using the same argument as above, as well as an upper bound.

\subsubsection{Problem Formulation}
Algorithm \ref{2layer_optimizer} shows how the weights of the 2-layers can be optimized independently, with the output from the first being an input to the second.
Recall that the error threshold $\bar{\phi}_Z$ estimated in (\ref{lower_bound}) provides only a lower bound to the maximum tolerable error. Thus, using this estimate may not yield the best approximation 
possible with the given $\phi_W$ and it is necessary to solve with higher values of the threshold $\phi_Z$ and look for better solutions (further aprroximations). We use a search-based method for this where we start with  $\phi_Z = \bar{\phi}_Z$ and keep increasing $\phi_Z$ in steps as long as it's not large enough to make the optimization problem of the output layer (line \ref{problem2} of Algo \ref{2layer_optimizer}) infeasible, and then reduce it to within a desired accuracy.
\begin{algorithm}[t]
\small
\caption{\small Problem Formulation for 2-layer NN}
\label{2layer_optimizer}
\begin{algorithmic}[1]
\State $h^r = $\Call{OptimWeights}{$I$,$L$,$\hat{Z}$, $\hat{x}$, $\hat{h}$, $M_Z$,$\phi_Z$}  \hfill $\forall$  $ r = 1 \ldots D$
%\Comment{Call this Problem 1}
%\State $h^r = tanh(b^r)$  \hfill  \dots $\forall$  $ r = 1 \ldots D$
\State $t^r =$ \Call{OptimWeights}{$L$,$J$,$\hat{W}$, ${h}$, $\hat{t}$, $M_W$,$\phi_W$}  \hfill $\forall$  $ r = 1 \ldots D$ \label{problem2}
%\Comment{Call this Problem 2}
%\State $t^r = tanh(a^r)$ \hfill  \dots $\forall$  $ r = 1 \ldots D$
\end{algorithmic}
\end{algorithm}
%Algorithm \ref{binary_search} provides a pseudo-code which repeatedly solves Algorithm \ref{2layer_optimizer} with diifferent values of $\phi_Z$ (but starting from $\bar{\phi}_Z$) until it is accurate to within some $\epsilon$.
%\begin{algorithm}[!t]
%\caption{Binary search for accurate $\phi_Z$}
%\label{binary_search}
%\begin{algorithmic}[1]
%\State $\phi_Z = \bar{\phi}_Z$
%\State $step = \bar{\phi}_Z$
%\While{$step \geq \epsilon$}
%    \State Solve Algorithm \ref{2layer_optimizer} with $\phi_Z$
%    \If {(Problem 2 (line \ref{problem2} of Algo \ref{2layer_optimizer}) was feasible)}
%	\State  $\phi_Z = \phi_Z + step$
%    \Else
%	\State $step = step/2$
%	\State $\phi_Z = \phi_Z - step$
%    \EndIf
%\EndWhile
%\end{algorithmic}
%\end{algorithm}
\setlength{\intextsep}{6pt plus 1.0pt minus 2.0pt}
\section{Experimental Methodology and Results}
\label{results}
Several benchmarks based on classification problems were used to measure the performance of the NNs and estimate the energy savings obtained by approximating the multiplications. First, we train an NN in MATLAB using the error back-propagation method, check its accuracy on the test dataset and estimate its power with the Normal MTJ-SNG. Next, we incorporate BMS (introduced in sec. \ref{BMS}) and approximate the network using the optimization technique described in the previous section for different levels of error tolerance. For solving the optimization problems we use CVX, a package for specifying and solving convex programs \cite{cvx}. Finally, each of the newly obtained NNs with approximate multipliers were analyzed for their accuracy and power. 

The power consumption of the Normal MTJ-SNG and BMS in the networks were obtained from the data corresponding to the red and green (dotted) plots respectively in fig. \ref{bms}\subref{bms_energy} and those of the FSM-based $tanh$ from \cite{ardakani2016vlsi}. The results from the different datasets are summarized below:

\textbf{MNIST digit recognition:}
The MNIST is a standard benchmark for classification problems that categorizes handwritten digits, each of size $28 \times 28$ \cite{lecun1998mnist}. First, a simple 1-layer NN with 784 inputs and 10 outputs was trained. This original network had an accuracy of $87.43\%$ on the test dataset and power consumption of $707.9mW$ with the Normal MTJ-SNG. Table \ref{mnist_1layer} shows the benefit of replacing that with BMS. As can be seen from the $1^{st}$ row, just the use of BMS reduces the power to $545.1mW$, while maintaining the same accuracy. The following rows summarize the optimization results for different values of error tolerance. Significant energy  savings were obtained even for $\phi_W = 0$ owing to certain degree of redundancy in some inputs. Classification accuracy drops by only 1.4\% (with $\phi_W = 0.03$) for about 39\% reduction in power when compared to the original, with 21\% reduction over BMS ($1^{st}$ row) being achieved through optimization. We also observe how the classification accuracy, as well as the computation energy, vary with the precision of the stochastic streams.  

\small
\begin{table}[h]
\centering
\caption{\footnotesize Variation of power with error, and of accuracy \& energy with both error and precision for the MNIST 1-layer network with BMS. The \textit{Inf} column corresponds to a theoretically infinite precision and the $1^{st}$ row corresponds to BMS without any weight approximation. }
\label{mnist_1layer}
\setlength{\tabcolsep}{2.5pt}
\renewcommand{\arraystretch}{1.2}
\begin{tabular}{c c|c|c|c|c||c|c|c|}
\cline{3-9}
&   & \multicolumn{4}{c||}{Accuracy (in \%)} & \multicolumn{3}{c|}{Energy$(\mu J)$} \\
%\cline{1}
%\cline{1-1} \cline{3-9}
\hline
\multicolumn{1}{|c|}{$\phi_W$} & \diagbox[height=0.45cm, innerleftsep=0cm,innerrightsep=0pt]{Power}{Precision}  &Inf & 512 & 256 & 128 & 512 & 256 & 128 \\
%\hline
%\multicolumn{1}{|c|}{--} & 708 & 87.43 & 87.41 & 87.59 & 81.97 & 3762 & 1881 & 941\\
\hline
\multicolumn{1}{|c|}{--} & $545.1$ $mW$ & 87.43 & 87.41 & 87.59 & 81.97 & 2.18 & 1.09 & 0.54\\
\hline
\multicolumn{1}{|c|}{0.00} & $443.8$ $mW$ & 87.10 & 87.05 & 87.26 & 81.67 & 1.77 & 0.88 & 0.44\\
\hline
\multicolumn{1}{|c|}{0.01} & $435.2$ $mW$ & 86.74 & 86.79 & 86.59 & 83.51 & 1.74 & 0.87 & 0.43\\
\hline
\multicolumn{1}{|c|}{0.02} & $431.4$ $mW$ & 86.57 & 85.95 & 86.42 & 82.43 & 1.72 & 0.86 & 0.43\\
\hline
\multicolumn{1}{|c|}{0.03} & $\textbf{428.2}$ $\textbf{\textit{mW}}$ & \textbf{86.03} & 86.17 & 85.82 & 84.29 & 1.71 & 0.85 & 0.42\\
\hline
\end{tabular}
\end{table}
\normalsize
For the 2-layer NN, input images were scaled down to size $14 \times 14$ to reduce the complexity of the problem (and hence the time required to solve it), and 15 neurons were used in the hidden layer. The original network had an accuracy of 92.28\% and power of $273.68mW$. Results of incorporating BMS and then approximating the network are shown in Table \ref{mnist_2layer}, using $\phi_Z$ values that gave the least power. From the last column, it is evident that 43\% decrement in power over the original, and 26\% over BMS without approximation, was obtained with less that 1\% degradation of accuracy.

\begin{table}[h]
\centering
\caption{\footnotesize MNIST 2-layer result. $1^{st}$ column is BMS w/o approximation} % Variation of power \& accuracy with $\phi_W$
\label{mnist_2layer}
\setlength{\tabcolsep}{1.2pt}
\renewcommand{\arraystretch}{1.2}
\begin{tabular}{|c|c|c|c|c|c|c|c|c|}
\hline
$\phi_W$ & -- & 0.00 & 0.01 & 0.02 & 0.03 & 0.04 & 0.06 & 0.08 \\
\hline
$\phi_Z (\times 10^{-2})$ & -- & 0.00 & 0.64 & 1.28 & 1.92 & 2.56 & 2.88 & 3.84 \\
\hline
Accuracy(in\%) & 92.28 & 92.18 & 92.14 & 92.07 & 92.02 & 91.97 & 91.72 & \textbf{91.35} \\
\hline
Power$(mW)$ & 210.74 & 175.48 & 169.26 & 165.06 & 161.94 & 159.59 & 158.36 & \textbf{155.46} \\
\hline
\end{tabular}
\end{table}

\textbf{SONAR, Rocks vs. Mines: }
This dataset (as well as the next one) was obtained form the UCI Machine Learning Repository \cite{Lichman:2013}. It requires us to distinguish between metal surfaces and rocks using sonar signals bounced off from them \cite{gorman1988analysis}. Both the training and test datasets contain 104 samples, each having 60 inputs. The accuracy and power of the original networks were 83 and $11.49mW$ respectively for the 1-layer NN, and 90 and $62.27mW$ for the 2-layer. Effect of using BMS and performance of our algorithm on both 1-layer NN and 2-layer NN (with 12 hidden neurons) are in Table \ref{sonar}. Unlike MNIST, here there was no power savings for $\phi_W=0$. For the 1-layer NN, we observe 60\% power reduction over original with accuracy degradation of 5 (for $\phi_W=0.2$). For the 2-layer NN, the respective figures are 57\% and 3 respectively.
%The $1^{st}$ column lists those of the original trained networks .

%\begin{table}[h]
%\caption{rfr}
%\label{sonar_1layer}
%\setlength{\tabcolsep}{2.0pt}
%\renewcommand{\arraystretch}{1.1}
%\begin{tabular}{|c|c|c|c|c|c|c|c|}
%\hline
%$\phi_W$ & 0.01 & 0.02 & 0.03 & 0.04 & 0.05 & 0.06 & 0.07 \\
%\hline
%Accuracy & 83 & 83 & 82 & 82 & 82 & 81 & 78 \\
%\hline
%Power$(mW)$ & 7.71 & 7.18 & 6.81 & 6.63 & 6.48 & 6.30 & 6.14 \\
%\hline
%\end{tabular}
%\end{table}

%\begin{table}[h]
%\centering
%\caption{Results for SONAR 2-layer.} %Variation of power \& accuracy with $\phi_W$
%\label{sonar_2layer}
%\setlength{\tabcolsep}{2.0pt}
%\renewcommand{\arraystretch}{1.1}
%\begin{tabular}{|c|c|c|c|c|c|c|}
%\hline
%$\phi_W (\times 10^{-2})$ & 1 & 2 & 5 & 10 & 15 & 20 \\
%\hline
%$\phi_Z (\times 10^{-2})$ & 0.675 & 1.48 & 3.9 & 7.8 & 14.625 & 18.2 \\
%\hline
%Accuracy & 90 & 89 & 89 & 91 & 90 & 87 \\
%\hline
%Power$(mW)$ & 44.05 & 40.96 & 36.24 & 33.03 & 29.19 & 27.02 \\
%\hline
%\end{tabular}
%\end{table}

\begin{table}[h]
\centering
\caption{\footnotesize SONAR 1 and 2-layer. $1^{st}$ column is BMS w/o approximation} %Variation of power \& accuracy with $\phi_W$
\label{sonar}
\setlength{\tabcolsep}{2.0pt}
\renewcommand{\arraystretch}{1.2}
\begin{tabular}{c|c|c|c|c|c|c|c|c|}
\cline{2-9}
& $\phi_W (\times 10^{-2})$ & -- & 1 & 2 & 5 & 10 & 15 & 20 \\
\hline
\multicolumn{1}{ |c|  }{\multirow{2}{*}{1-layer NN}} & Accuracy & 83 & 83 & 83 & 82 & 81 & 81 & \textbf{78} \\
\cline{2-9}
\multicolumn{1}{ |c|  }{} & Power$(mW)$ & 8.85 & 7.71 & 7.18 & 6.48 & 5.66 & 4.96 & \textbf{4.62} \\
\hline
\multicolumn{1}{ |c|  }{\multirow{3}{*}{2-layer NN}} & $\phi_Z (\times 10^{-2})$ & -- & 0.675 & 1.48 & 3.9 & 7.8 & 14.625 & 18.2 \\
\cline{2-9}
\multicolumn{1}{ |c|  }{} & Accuracy & 90 & 90 & 89 & 89 & 91 & 90 & \textbf{87} \\
\cline{2-9}
\multicolumn{1}{ |c|  }{} & Power$(mW)$ & 47.95 & 44.05 & 40.96 & 36.24 & 33.03 & 29.19 & \textbf{27.02} \\
\hline
\end{tabular}
\end{table}

\textbf{Wine Quality: }
The goal is to train a network to estimate the quality of samples (on a scale of 10) of red and white wine on the basis of results of physiochemical tests \cite{Cortez2009547}. 
%There are a total of 1599 and 4898 samples of red and white wine respectively. 
Only 2-layer NNs with 20 hidden neurons were trained that gave accuracy of 86.4\% and power $30.56mW$ with red wine, and 85.75\% and $32.83mW$ respectively with white wine. Results are displayed in Table \ref{wine_quality}. Power savings over the original are 38\% for accuracy loss of 1.2\% for Red wine, and 42\% with loss in accuracy of 0.45\% for White wine.

\small
\begin{table}[h]
\centering
\caption{\footnotesize Wine Quality - The $1^{st}$ row is BMS without approximation}
\label{wine_quality}
\setlength{\tabcolsep}{1.5pt}
\renewcommand{\arraystretch}{1.2}
\begin{tabular}{c|c|c|c||c|c|c|}
\cline{2-7}
& \multicolumn{3}{c||}{Red Wine} & \multicolumn{3}{c|}{White Wine}  \\
\hline
\multicolumn{1}{|c|}{$\phi_W$} & $\phi_Z$ & Accuracy & Power$(mW)$ & $\phi_Z(\times 10^{-2})$ & Accuracy & Power$(mW)$  \\
\hline
\multicolumn{1}{|c|}{--} & -- & 86.4\% & 23.53 & -- & 85.75\% & 25.28  \\
\hline
\multicolumn{1}{|c|}{0.02} & 0.021 & 86.4\% & 22.40 & 4.32 & 86.09\% & 22.39 \\
\hline
\multicolumn{1}{|c|}{0.04} & 0.042 & 86.0\% & 21.36 & 2.16 & 85.86\% & 21.76 \\
\hline
\multicolumn{1}{|c|}{0.10} & 0.042 & 86.4\% & 20.14 & 2.16 & 86.86\% & 20.46 \\
\hline
\multicolumn{1}{|c|}{0.15} & 0.056 & 85.6\% & 19.30 & 6.48 & 85.97\% & 19.79 \\
\hline
\multicolumn{1}{|c|}{0.20} & 0.070 & \textbf{85.2\%} & \textbf{18.89} & 2.16 & \textbf{85.30\%} & \textbf{19.19} \\
\hline
%\multicolumn{1}{|c|}{0.12} & 0.112 & 14.8\% & 20.28 & 8.64 & 14.58\% & 20.20 \\
%\hline
\end{tabular}
\end{table}
\normalsize

\section{Conclusion}
\label{conclusion}
This paper proposes the use of MTJs as SNGs in an SC based hardware implementation of Neural Networks. We design a low-power version of an MTJ-SNG (named BMS) that significantly reduces the average energy per bit of a stochastic stream and propose its use in an SC-based NN. We go on to develop an algorithm based on convex optimization that adjusts the weights in such an NN by leveraging the error resilient nature of applications of NNs. Classification problems were evaluated on this approximate NN and results showed substantial gains in energy savings for little loss in accuracy.

\section{Acknowledgement}
\label{acknow}
This work is supported by the Air Force Office of Scientific Research under Grant FA9550-14-1-0351.

\bibliographystyle{IEEEtran}
%\bibliography{References}
\bibliography{ISLPED_paper_arXiv_2}

\end{document}